%% file: main.tex
\documentclass[letterpaper, 10 pt, conference]{ieeeconf}  

\IEEEoverridecommandlockouts                              

\usepackage{multicol}
\usepackage{multirow}
\usepackage[bookmarks=true]{hyperref}
\usepackage{amssymb}
\usepackage{tabularx}
\usepackage{amsmath, mathtools, cuted}
\usepackage{algorithm2e}
\usepackage{caption}
\usepackage{subcaption}
\usepackage{float}
\usepackage{fancyhdr}
\usepackage[backend=bibtex,style=ieee]{biblatex}
\bibliography{references.bib}

\title{\LARGE \bf
Learning Neuro-symbolic Programs for Language Guided \\ Robot Manipulation}

\author{
Namasivayam K$^{*}$$^{1}$, 
Himanshu Singh$^{*}$$^{1}$, 
Vishal Bindal$^{*}$$^{1}$, 
Arnav Tuli$^{1}$, 
Vishwajeet Agrawal$^{\#}$$^{2}$, 
Rahul Jain$^{\#}$$^{2}$, \\
Parag Singla$^{1}$ and Rohan Paul$^{1}$\\
\small $^{1}$Affilitated with IIT Delhi. 
\small $^{2}$Work done when at IIT Delhi.
$^{*}$ and ${^{\#}}$ denote equal contribution.
}

\newcommand\real{\mathbb{R}}
\usepackage{contour}
\usepackage{graphicx}
\usepackage{bbm}
\contourlength{0.8pt}

\begin{document}

\maketitle
\thispagestyle{fancy}%
\pagestyle{empty}

\input{abstract}

\input{intro}

\input{related}

\input{problem}
\input{approach}
\input{experiment2}
\input{conclusions}
\input{ack}
\newpage
\printbibliography

\end{document}

%% file: abstract.tex
\begin{abstract}
%
%
Given a natural language instruction and an input scene, our goal is to train a model 
to output a \emph{manipulation program} that can be executed by the robot.
%
%
Prior approaches for this task possess one of the following limitations: (i) rely on hand-coded symbols for concepts limiting generalization beyond those seen during training~\cite{paul2016efficient} (ii) infer action sequences from instructions but require dense sub-goal supervision~\cite{paxton2019prospection} or (iii) 
lack semantics required for deeper object-centric reasoning inherent in interpreting complex instructions~\cite{shridhar2022cliport}.
%
%
%
In contrast, our approach can handle linguistic as well as perceptual variations, end-to-end trainable and requires no intermediate supervision. The proposed model uses symbolic reasoning constructs that operate on a latent neural object-centric representation, allowing for deeper reasoning over the input scene.
%
Central to our approach is a modular structure  consisting of a {\em hierarchical instruction parser} and an {\em  action simulator} to learn  disentangled action representations.
%
%
%
Our experiments on a simulated environment with a 7-DOF manipulator, consisting of instructions with varying number of steps and scenes with different number of objects,  demonstrate that our model is robust to such variations and significantly outperforms baselines, particularly in the generalization settings. The code, dataset and experiment videos are available at \href{https://nsrmp.github.io/}{\texttt{https://nsrmp.github.io}} 
\end{abstract}

%% file: intro.tex
\section{Introduction}
\label{sec:intro}
\begin{figure*}
    \centering    
    \includegraphics[width=16cm]{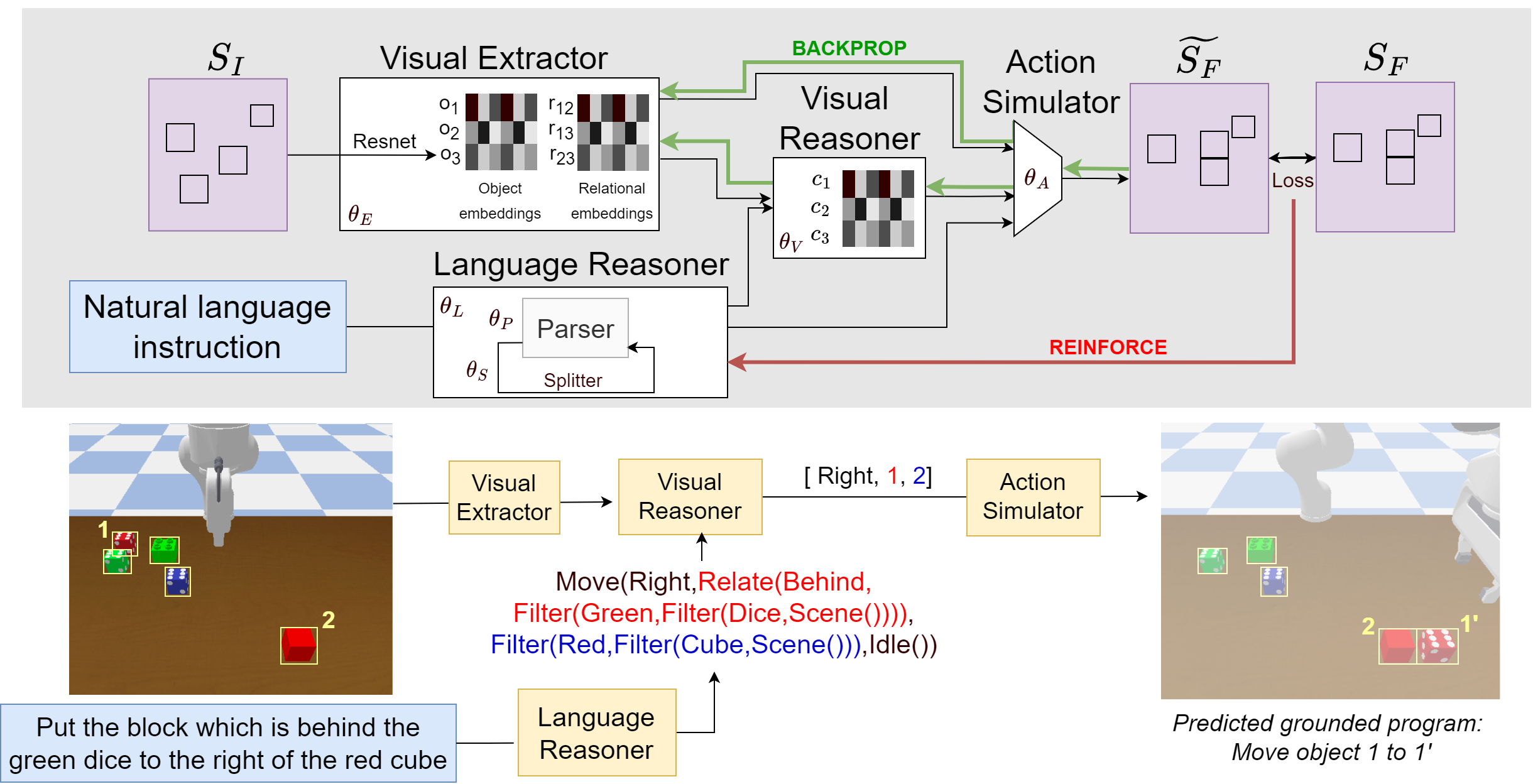}
    
    \caption{
    \footnotesize{
    \textbf{Model architecture.} 
    The \emph{Visual Extractor} forms dense object representations from the scene image using pre-trained object detector and feature extractor.  
    The \emph{Language Reasoner} auto-regressively induces a symbolic program  from the instruction that represents rich symbolic reasoning over spatial and action constructs inherent in the instruction.  
    The \emph{Visual Reasoner} determines which objects are affected by actions in the plan using symbolic and spatial reasoning.  
    The \emph{Action Simulator} predicts final location of the moved object. 
    The model is trained end-to-end with a loss on the bounding boxes, 
    %
    backpropagated to action and visual modules. REINFORCE is used to train \cite{williams1987class} the language reasoner from which symbolic programs are sampled.     }
    \vspace{-0.5cm}}
    
    \label{fig:schematic}
\end{figure*}

As robots enter human-centric environments, they must learn to act and achieve the intended goal based on natural language instructions from a human partner. 
We address the problem of learning to translate high level language instructions into
executable programs grounded in the robot’s state and action space. 
We focus on multi-step manipulation tasks that involve object interactions such as stacking and assembling objects referred to by their attributes and spatial relations. 
Figure~\ref{fig:schematic} provides an example where the robot is instructed to 
\emph{``put the block which is behind the green dice to the right of the red cube"}.  
We assume the presence of natural supervision from a human teacher in the form of 
input and output scenes, along with linguistic description for a high-level manipulation task. Our goal 
is to learn representations for visual and action concepts that can be composed to achieve the task. 
The learning problem is hard since (1) object attributes and actions have to be parsed from the underlying sentence (2) object references need to be grounded given the image and (3) the effect of executing the specified actions has to be deciphered in the image space, requiring complex natural language as well as image level reasoning. Further, the model needs to be trained end-to-end, learning representations for concepts via distant supervision.

Prior efforts can be broadly categorized as 
(i) Traditional methods which learn a mapping between phrases in the natural language to symbols representing robot state and actions in a pre-annotated dataset~\cite{howard2014natural,paul2016efficient,tellex2011approaching,matuszek2013learning,knepper2013ikeabot,gopalan2018sequence,williams2018learning}; they lack the flexibility to learn the semantics of concepts and actions on their own, an important aspect required for generalizability (ii) Approaches that model an instruction as a sequence of action labels to be executed, without any deeper semantics, and requiring intermediate supervision for sub-goals, which may not be always be available~\cite{paxton2019prospection,shah2018bayesian,wang2020learning,kress2008translating,lazaro2019beyond,tenorth2010understanding,lisca2015towards,misra2016tell} 
(iii) Recent end-to-end learning approaches~\cite{konidaris2018skills,wang2021learning, zettlemoyer2005learning,xia2018learning,silver2020few,zhu2021hierarchical,shridhar2022cliport,zeng2020transporter}; they have limited reasoning capability both at the level of instruction parsing, and their ability to learn varied action semantics.

In response, we introduce a neuro-symbolic approach for jointly learning representations for concepts that can be composed to form \emph{manipulation programs}: symbolic programs that explain how the world scene is likely to affected by the input instruction. Our approach makes use of a Domain Specific Language (DSL) which specifies various concepts whose semantics are learned by the model. 
We build on the concept learning framework by \cite{Mao2019NeuroSymbolic} for grounding of concepts in a natural language instruction in the image. The \emph{manipulation program} is grounded into the scene to get a sequence of actions that are fed to the low-level motion planner of the robot, which is assumed to be given.



The contributions of this work are: 
(i) A novel neuro-symbolic model that learns 
to perform complex object manipulation tasks requiring reasoning over scenes for a natural language instruction, given initial 
and final world scenes.  
(ii) A demonstration of how dense representations for robot manipulation actions 
can be acquired using only the initial and final world states (scenes) as supervision without the need for any intermediate supervision. 
(iii) An RL-based learning approach that facilitates parsing of the complex instruction into symbolic concepts, and allows for learning of disentangled actions in the robot's space. 
(iv) Evaluation in instruction following experiments with a simulated 7-DOF robot manipulator, demonstrating robust generalization to novel settings improving on the state-of-the-art.

%% file: related.tex
\section{Related Works}

\textbf{Grounding Instructions to Robot Control.} Traditional approaches for grounding instructions assume a symbolic description of the environment state 
and robot actions. 
These approaches learn to associate phrases in an instruction with symbols conveying their meaning from 
annotated datasets. 
For example, \cite{howard2014natural,paul2016efficient,tellex2011approaching} map language to discrete motion constraints that can be provided to a motion planner for trajectory generation, \cite{matuszek2013learning,knepper2013ikeabot} parse language to a logical parse that includes  robot control actions and \cite{gopalan2018sequence,williams2018learning} use sequence to sequence modeling to ground language to LTL formulae capturing high-level task specifications. 
%
%
In contrast to such approaches, the proposed work doesn't assume a symbolic model of the world and learns spatial and visual concepts and action semantics from data. 

\textbf{Inferring Plans from Instructions.} Another set of approaches focus on multi-stage tasks (e.g., cooking, assembly etc.) and propose 
learners that can infer action sequences by observing human task demonstrations. 
The work in ~\cite{paxton2019prospection} proposes a recurrent neural architecture that converts a natural language instruction to a sequence of intermediate goals for robot execution. 
Efforts such as \cite{shah2018bayesian,wang2020learning,kress2008translating} learn LTL task specifications from humans demonstrations and introduce a model for effectively searching in the space of programs. 
In another effort, ~\cite{lazaro2019beyond}, authors introduce a cognitively-inspired approach that infers 
a likely program from a symbolic generative grammar for a robot manipulator to form and re-arrange block patterns
in a table top setting. 
%
%
Misra et al. \cite{misra2016tell} present a similar approach but additionally assess plan feasibility for inferred plan candidates 
using a symbolic planner in the training loop. 
Despite successful demonstrations, these approaches require dense  
intermediate supervision and treat robot actions in a purely symbolic manner without learning their deep grounded semantics. 
%
However, our method learns a latent space of composable symbolic programs which can be executed in the latent space of object representations. The composability of the  symbolic programs enables incremental learning through a specified curriculum from only the initial and final state data, forgoing the need for dense intermediate supervision.

\textbf{Learning Action Models.} This work builds on the rich literature on acquiring action models for planning and captures \emph{when} actions are applicable and 
\emph{how} they affect the world. 
Efforts such as \cite{konidaris2018skills} and \cite{wang2021learning} 
learn initiation sets for actions implicitly learning to classify
the robot's configuration space where an action can be initiated. 
%
%
~\cite{xia2018learning,silver2020few,zhu2021hierarchical} present neuro-symbolic approaches for learning a latent object-centric world model for tasks such as stacking, re-arrangement etc.  
In \cite{shridhar2022cliport}, authors introduce an end-to-end model that jointly predicts object affordances  and object displacements specifying metric goals for robot re-arrangement and sorting tasks. 
Recent parallel work by Wang et al. \cite{wang2023programmatically} builds on \cite{shridhar2022cliport} and represents the task as an executable program parsed from the instruction using a CCG parser, restricting the length and variation of the instructions it can handle. 
Even though these approaches successfully learn modular action models that are amenable to planning, the scope of reasoning is limited to one-hop reasoning over spatial relations and attributes. In contrast, this work focuses on interpreting task instructions that require compositional/hierarchical spatial reasoning over an extended planning horizon.

%% file: problem.tex
\section{Problem Formulation}\label{sec:problem}
The robot perceives the world state comprising a set of rigid objects placed on a table via a depth sensor 
that outputs a depth image $S \in \real^{H\times W \times C}$, where $H,W,C$ respectively
denote the height, width and the number of channels (including depth) of the imaging sensor.  
%
%
%
The workspace is co-habited by a human partner 
who provides language instructions 
%
to the robot to perform assembly tasks.
We assume a closed world setting, i.e. changes to the world state are caused only by the robot manipulator. 

The robot's goal is to interpret the human's instruction $\Lambda$ 
in the context of the initial world state $S_I$ and determine a 
sequence of low-level  motions that result in the final world state 
$S_F$ conforming to the human's intention.  
%
%
Following ~\cite{kaelbling2010hierarchical, zhu2020hierarchical}, 
planning for a complex task is factorized into  (i) high-level task planning to determine a sequence of sub-goals, and (ii) the generation of 
low-level motions to attain each sub-goal. 
Formally, a semantic model for a manipulation task denoted as $ManipulationProgram(.)$ takes the initial scene $S_I$ and the instruction $\Lambda$ as input and determines a sequence of 
sub-goals as $(g_0, g_1, \dots, g_n) = ManipulationProgram(S_I, \Lambda)$. 
Each sub-goal $g_i$ aggregates the 
knowledge of the object, $o_i$, to be manipulated 
and its target Cartesian $SE(3)$ pose $p_i$. This is  
provided to the low-level motion planner 
to synthesize the end-effector trajectory for the robot to execute. 
The robot's motion planner, which is assumed to be given, includes grasping an object  
and synthesizing a collision-free trajectory for positioning it at a target pose. 
%
%

%
%
Training data consists of demonstrations, corresponding to task instructions $\Lambda$, in the form of initial 
and final 
world states $(S_I, S_F)$.
Given data $D$$=$$\{ S^{i}_I, S^{i}_F, \Lambda^{i} \}^{M}_{i=1}$, the model parameters are trained by optimizing a loss 
$\mathcal{L}(\tilde{S}_F^{i},S_F^{i})$
%
where $\tilde{S}_F^{i}$$=$$Simulate(ManipulationProgram(S_I^{i}, \Lambda; \Theta))$ is the final state estimated by simulating the plan inferred by $ManipulationProgram(S_I^{i}, \Lambda)$ on initial state $S_I^i$.  
%
We seek strong generalization on novel scenes, instructions and plan lengths 
beyond those encountered during training, along with interpretability in sub-goals. 

%% file: approach.tex
\section{Technical Approach}
\begin{figure*}[htbp]
  \centering
  \begin{subfigure}[b]{0.4\textwidth}
    \includegraphics[width=\textwidth]{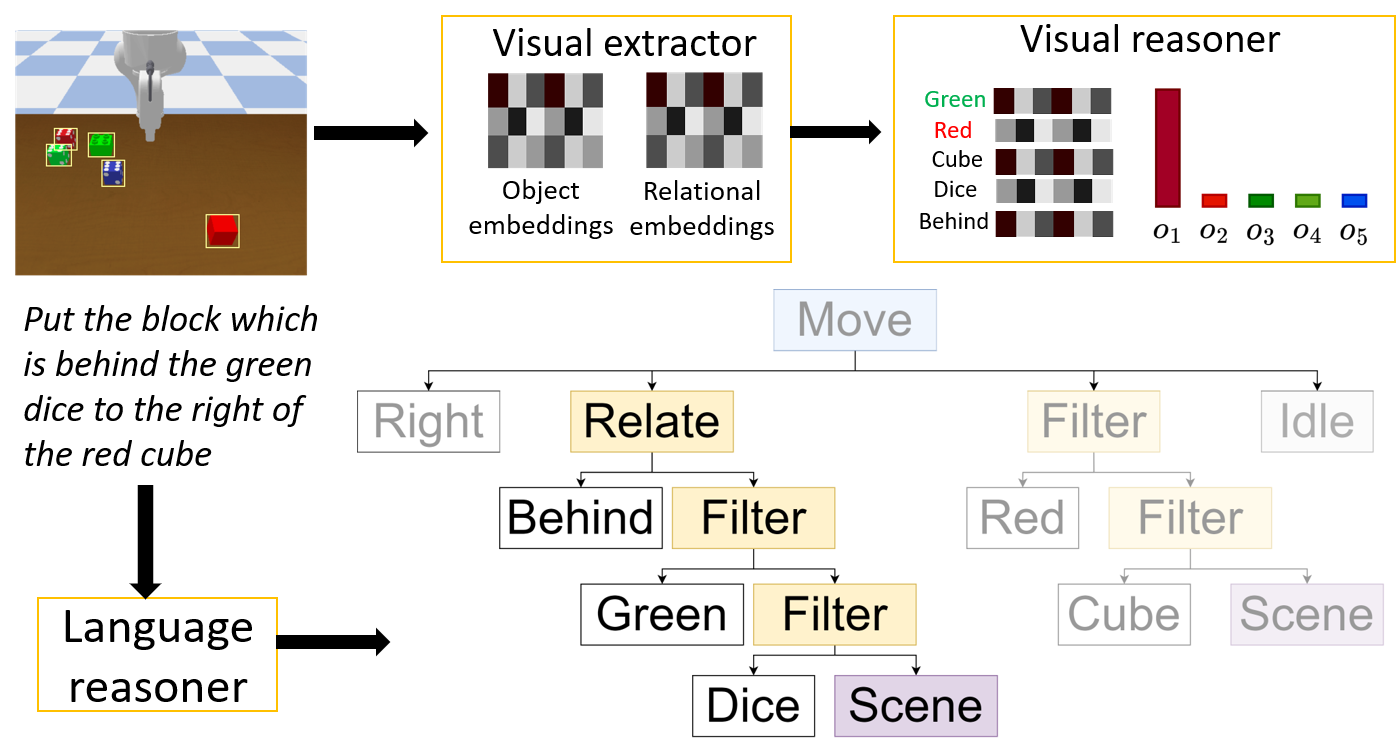}
    \caption{}
    \label{fig:approach2}
  \end{subfigure}\hfill
  \begin{subfigure}[b]{0.5\textwidth}
    \includegraphics[width=\textwidth]{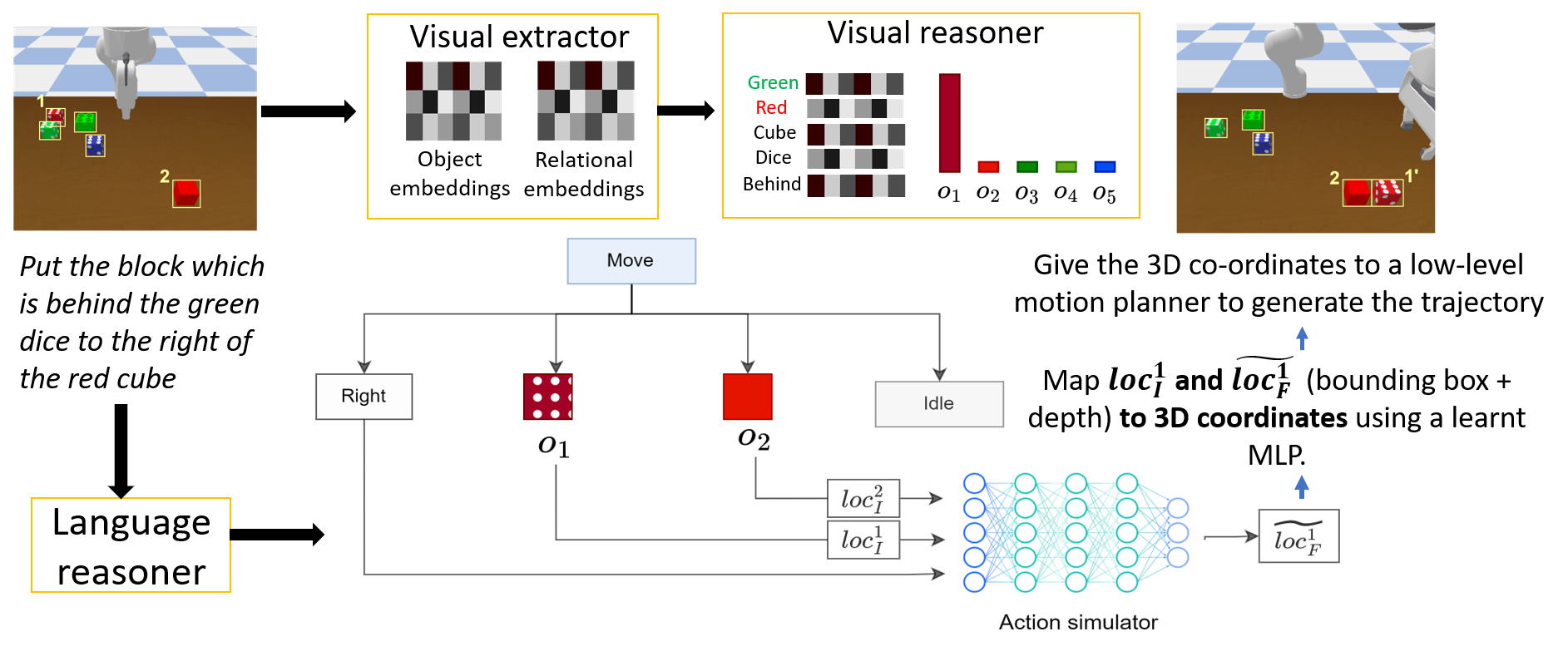}
          \caption{}
    \label{fig:approach1}
  \end{subfigure}
   
  \caption{\footnotesize{\textbf{Quasi-symbolic program execution.} We postulate a latent space of hierarchical, symbolic programs that performs explicit reasoning over action, spatial and visual concepts. a) The language reasoner infers a program belonging to this space, capturing the semantics of the language instruction. The program is executed over the latent object representations extracted from the initial scene to get a grounded program. b) The grounded program is used by the action simulator to compute the final location of the moved object. This is fed into a low-level motion planner for computing the low-level trajectory.}}
  \label{fig:approach}
  \vspace{-0.1 in}
\end{figure*}

We propose a neuro-symbolic architecture to solve the task planning problem described in section \ref{sec:problem}. Our architecture is inspired by ~\cite{Mao2019NeuroSymbolic} and is trained end-to-end with no intermediate supervision.
We assume that the reasoning required to infer the sub-goals can be represented as a program determined by a domain specific language (DSL). Table \ref{table:dsl} lists the keywords and operators in our DSL, along with the implementation details of the operators. We assume a lexer that identifies all the keywords that are referred to in the  instruction $\Lambda$. We do not assume prior knowledge of the semantics of the DSL constructs, and they are learned purely from the data.
 Our architecture (ref. Figure~\ref{fig:schematic}) consists of the following key modules (a) Language Reasoner (LR) to parse the instruction into a hierarchical plan consisting of references to visual and action concepts and operators (b) Visual Extractor (VE) to obtain the initial bounding boxes from the input image (c) Visual Reasoner (VR) to ground the object related concepts present in the parsed instruction with respect to the input image (d) Action Simulator (AS) to learn the semantics of actions to be executed to produce a sequence of sub-goals. Figure ~\ref{fig:approach} illustrates our approach. 


\begin{table}
    \begin{tabular}{|p{0.01cm}|p{0.01cm}|p{0.08cm}|p{0.08cm}|}
        \hline
        \multicolumn{4}{|c|}{\textbf{Keywords and their classes}}\\
        \hline
        
         \multicolumn{2}{|c|}{\textbf{Object-level concepts}}& \multicolumn{2}{c|}{\textbf{Other concepts}} \\
         \multicolumn{2}{|l|}{\textbf{Color}: \{Red, Blue, Cyan,...\}}& \multicolumn{2}{l|}{\textbf{RelCpt}: \{Left, Behind, Front,...\}} \\
         \multicolumn{2}{|l|}{\textbf{Type}: \{Cube, Lego, Dice\}} & \multicolumn{2}{c|}{\textbf{ActCpt}:  \{MovRight, MovTop,...\}} \\
    \hline
    \hline
    \multicolumn{4}{|c|}{\textbf{Operators: ( Input $\rightarrow$ Output)}}\\
    \hline
    \multicolumn{2}{|l|}{Scene : None $\rightarrow$ ObjSet} & \multicolumn{2}{l|}{Unique: ObjSet $\rightarrow$ Obj}\\
     \multicolumn{2}{|l|}{Filter : (ObjSet, ObjCpt)$ \rightarrow$ ObjSet} & \multicolumn{2}{l|}{Relate  : (Obj, RelCpt) $\rightarrow$ ObjSet}\\
     \multicolumn{2}{|l|}{ Move : (ActCpt, World) $\rightarrow$ World} & \multicolumn{2}{l|}{ Idle :  World $\rightarrow$ World}\\
    \hline
    \multicolumn{4}{|c|}{ObjectSet $\in \mathbb{R}^{\text{N}}$, N = Num objects, Object = one-hot ObjectSet}\\
    \multicolumn{4}{|c|}{World = $\{(b_i,d)\}_{i=1}^\text{N} \in \mathbb{R}^5$, bounding boxes and depth for all objects }\\
    \hline
        \end{tabular}
    \caption{Domain Specific Language.}
    \label{table:dsl}
    \vspace{-0.5cm}
\end{table}
    
\subsection{Language Reasoner (LR)}
The language reasoner (LR) deduces a hierarchical symbolic program that corresponds to the 
manipulation task implied by the human's utterance to the robot. 
The deduced program consists of symbolic reasoning constructs that operate on neural concepts grounded in the state space of the scene and the action space of the robot. 
%
%
%
Since a high-level task instruction may imply a sequence of actions, we adopt a hierarchical model where an LSTM~\cite{lstm} network first splits the instruction into a sequence of sub-instructions, each corresponding to one grounded action. The semantic parse of each sub-instruction is then inferred using a hierarchical  \emph{seq2tree} architecture similar to~\cite{dong2016language,Mao2019NeuroSymbolic}. The sequence of programs thus generated are composed to produce the symbolic program for the original language instruction.
\subsection{Visual Extractor (VE)}
\label{subsec:visual-reason}
The visual extractor forms an object-centric view of the world by extracting object proposals in the form of bounding boxes for the objects (and class labels) 
using a pre-trained object detector~\cite{redmon2016you}. 
Following~\cite{Mao2019NeuroSymbolic}, dense object representations are obtained by passing the bounding boxes (after non-maximal suppression) through a feature extractor as~\cite{targ2016resnet}. 
The data association between object proposals is estimated greedily based on cosine similarity between the dense object features in the initial/final scenes. 

\subsection{Visual Reasoner (VR)}
The visual reasoner performs visuo-spatial, object-centric reasoning to compute a sequence of sub-goals grounded in the scene from the symbolic program.
E.g., the instruction \textit{``put the block which is behind the green dice"} requires the robot to perform reasoning to resolve the specific world object that conforms to the relations/attributes as being behind a dice with colour attribute of green. 

The visual reasoner parameterizes object-level and relational concepts in the DSL (Table ~\ref{table:dsl}) with neural embeddings. Similar to ~\cite{Mao2019NeuroSymbolic}, a differentiable framework is defined for the execution of operators such as \emph{Filter, Unique, Relate, Idle}.  \\ 
As in ~\cite{Mao2019NeuroSymbolic}, intermediate results of execution are represented in a probabilistic manner. For e.g., a set of objects is represented as a vector of probabilities where the $i$-th element represents the probability of the $i$-th object being an element of the set. \emph{Filter, Unique, Relate} etc. sequentially modify this vector to ultimately yield the referred object/set of objects.  \\ 
This execution framework is used to execute the symbolic program  on top of the visual features (extracted by the visual extractor) resulting in the grounding of the program, $\Pi$, on the world state.

%

\subsection{Action Simulator (AS)}
\label{subsec:action-simulator}
The action simulator learns the semantics of action concepts of the DSL. It takes as input, the action concept, the initial locations  of the object to be manipulated and the reference object respectively and outputs the target location of the former. The outputted target location of the object being manipulated serves as a sub-goal for the low-level motion planner. We determine the location of an object by the corners $b = (x_1,y_1,x_2,y_2)$ of the enclosing bounding box and the depth of the object from the camera face. 

\noindent  
Overall, the model can be summarized as follows:
\begin{itemize}
    \item $\mathtt{P} \leftarrow \mathtt{LR}(\Lambda; \theta_P, \theta_S)$, where $\mathtt{P}$ is a symbolic program composed of DSL constructs. 
    \item $\Pi \leftarrow \mathtt{VR}(\mathtt{P}, \mathtt{VE}(S_I;\theta_E);\theta_V)$. The visual reasoner then grounds $\texttt{P}$ and outputs $\Pi$, the grounded program. For example, in Figure \ref{fig:schematic}, red and blue subprograms are grounded to object 1 and 2 respectively.
    \item $G \leftarrow \text{AS}(\Pi; \theta_A)$. The action simulator then takes $\Pi$ and returns a sequence of sub-goals, $G = (g_0, g_1, ..., g_{n-1})$, for the motion planner.
\end{itemize}
Here, $\theta$'s are the parameters of the corresponding  module. 

\subsection{Loss Function and Model Training}
\label{subsec:loss&training}
Given a single-step instruction $\Lambda$, the parser predicts a symbolic program $\mathtt{P}$. The visual reasoner grounds $\mathtt{P}$ to predict a sequence of sub-goals. The action simulator computes the low-level action corresponding to each sub-goal. The sequence of low-level actions is executed on the initial scene $S_I$ to get the predicted final locations of the objects $\{\widetilde{loc}^i_F\}_{i=1}^N$. Let $\{{loc}^i_F\}_{i=1}^N$ be the true locations in the gold final state, $S_F$. As mentioned above $loc = (b,d)$, where $b$ is the corners of bounding box and $d$ is the depth. The loss function $L_{act}\coloneqq \sum_i^N\|\widetilde{loc}^i_F- loc^i_F\|
+ \beta (1-\text{IoU}(\widetilde{b}^i_F, b^i_F))$ is used to train the action simulator and the visual modules. 
Since there is no explicit supervision to the parser, we train the parser using the policy gradient algorithm REINFORCE with the reward set to $-L_{act}$. 
During initial training, an explicit expectation (subtracting the mean action loss as the baseline) is computed over all programs to inform the loss, ameliorating the variance issue arising in REINFORCE.
The Language Reasoner is trained using REINFORCE and the other modules using backpropagation.

The following training curriculum is adopted: (i) training on single step commands with reasoning involving individual object features only, 
(ii) the trained action simulator module is frozen and additional training on single step commands involving reasoning over spatial relations between objects is carried out. 
(iii) the sentence splitter in the Language Reasoner module is fine-tuned on multiple-step instructions. 
\subsection{Scene Reconstruction}
We additionally train a neural model to synthesize the scene corresponding to each sub-goal, given only the initial scene and predicted object locations. This enables us to visualise the scene modification without the need of execution by a robot manipulator along with providing interpretability to the model's latent program space.
The reconstruction architecture is adapted from \cite{dhamo2020_SIMSG}, where the scene graph is constructed with nodes having object features and bounding boxes. This graph is updated with predicted bounding boxes at each step, yielding generated scenes. Presence of initial and final scenes in our data means we can train in a supervised manner, unlike \cite{dhamo2020_SIMSG}.

%% file: experiment2.tex
\section{Evaluation and Results}\label{sec:evaluation}
%

\begin{table*}[ht]
    \centering
    \caption{Comparison between Proposed Model and NMN+  (BB: Bounding Boxes)}
    \begin{tabular}{|c|c|c|c|c|c|c|c|c|c|c|c|c|}
    \hline
         Model  & \multicolumn{3}{|c|}{Overall} &  \multicolumn{2}{|c|}{Single-step} & \multicolumn{2}{|c|}{Double step} & \multicolumn{2}{|c|}{Simple} & \multicolumn{2}{|c|}{Complex} \\ 
         \hline
         \hline
          & IOU & IOU-M & Program & IOU & IOU-M  & IOU & IOU-M  & IOU & IOU-M & IOU & IOU-M  \\
           &  &  & (Action/Subj/Pred) & &   &  & &  &  &  &   \\
          \hline
         NMN+ (gold BB) & 0.77 & 0.55 & --/0.81/0.76 & 0.80 & 0.56 &    0.71 & 0.52 &0.90 & 0.71 & 0.64 & 0.31\\ 
         \hline 
         Ours (gold BB)  & 0.87 & 0.72& 0.99/0.99/0.94 & 0.91 & 0.64 &   0.87 & 0.62 & 0.92 & 0.73 & 0.87 & 0.64 \\
        \hline 
         NMN+ (extracted BB) & 0.69 & 0.32& --/0.81/0.55 & 0.78 & 0.49 &  0.58 & 0.22 & 0.79 & 0.55 & 0.60 & 0.20\\ 
         \hline 
         Ours (extracted BB) & 0.76  & 0.41  & 0.99/0.74/0.76 &0.80 & 0.43  & 0.69 & 0.54  & 0.83 & 0.62 & 0.69 & 0.24\\
         \hline 
    \end{tabular}
    \label{tab:accuracy}
\end{table*}

\begin{table}[ht]
    \large
    \centering
    \caption{Comparison b/w Proposed Model and CLIPort}
    \resizebox{\columnwidth}{!}{%
    \begin{tabular}{|c|c|c|c|c|c|c|c|c|c|c|c|c|}
    \hline
         Model  & \multicolumn{2}{|c|}{Overall}  & \multicolumn{2}{|c|}{Simple} & \multicolumn{2}{|c|}{Complex} \\ 
         \hline
         \hline
          & Id. Acc. & Pl. Acc. & Id. Acc.  & Pl. Acc. & Id. Acc.  & Pl. Acc.   \\
          
          \hline
            CLIPort & 0.46 & 0.39 & 0.55 & 0.47 & 0.32 &    0.26 \\ 
         \hline 
                     CLIPort(5x) & 0.76 & 0.61 & 0.90 & 0.74 & 0.55 &    0.39 \\ 
                     \hline 
         Ours (gold BB)  & 0.94 & 0.93 & 1.0 & 1.0  & 0.83 & 0.81  \\
         \hline 
    \end{tabular}%
    }
    \label{tab:cliport}
\end{table}

\subsection{Experimental Setup}
\textbf{Data collection. } 
The dataset is collected in a PyBullet tabletop environment using a 
simulated Franka Emika Panda robot arm. The workspace consists of a tabletop with blocks of different shapes and colors.
Each datapoint consists of an initial scene paired with a language instruction and the expected resulting final scene. 
For training, close to 5000 synthetic scenes and language instructions(with a 80:20 train-test split) are sampled with 3-5 blocks of varying colors and shapes and placed at randomized orientations on the table. 
%
%

The dataset consists of both single-step(e.g. \emph{``Put the red block on the green block"}) and double step commands(e.g. \emph{``Put the red block which is behind the white dice to right of the blue cube, then put the lego block on top of red block"}). 
On the basis of the level of reasoning involved in the given instruction, we can classify the dataset into (i) \textit{simple} that involve reasoning over individual object features only and \textit{complex} that additionally involve reasoning over inter-object relationships.

\subsection{Baselines, Metrics and Comparisons} 
The proposed model (i) performs visuo-linguistic reasoning in a symbolic latent space and (ii) assumes an object-centric state representation. We provide comparisons with baselines that forego these two assumptions.

\begin{figure}
\begin{subfigure}{1.0\hsize}
     \centering    
     \includegraphics[scale=0.19]{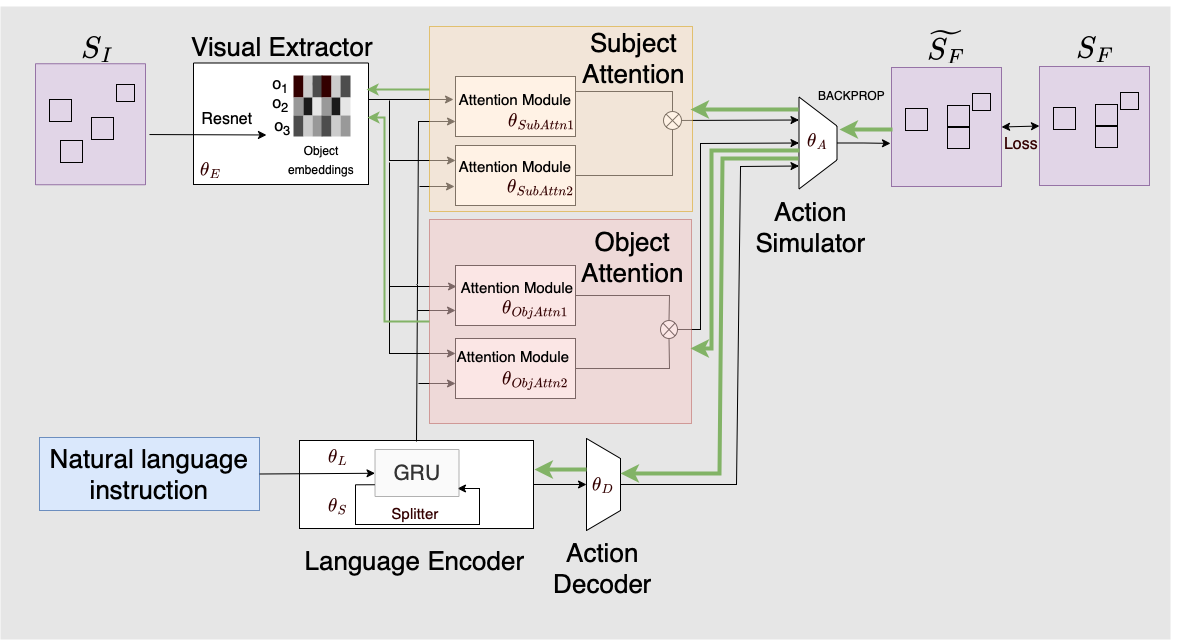}
\end{subfigure}
\caption{
    \footnotesize{
        \textbf{Object-centric baseline NMN+.} For a fair comparison, our model and NMN+ share the same language encoder (with LSTM-based splitter) and the visual extractor. Attention blocks compute language-guided attention over object embeddings to get the \emph{subject} and \emph{predicate} for the manipulation action. This is fed into the action simulator along with the action embedding from the action decoder to get the predicted final location of the object.}}
        \vspace{-0.15in}
            \label{fig:nmn}
\end{figure}

\textbf{NMN+}, inspired from Neural Module Networks \cite{andreas2016neural}, assumes an object-centric state representation. However, in contrast to the proposed model, visuo-linguistic reasoning required for manipulation is performed using language-guided attention over object embeddings instead of using symbolic reasoning constructs. Figure~\ref{fig:nmn} provides the detailed architecture. The entire architecture is neural and is trained end-to-end with the loss same as that of the proposed model. 


The following metrics are used for comparing the proposed model with \emph{NMN+}:
(i) \emph{Intersection over Union (IOU)} of the predicted and groundtruth bounding box is calculated in the 2D image space assuming a static camera viewing the scene. Average IOU over all objects in the scene and mean IOU for objects moved during execution, termed IOU-M,  are reported.  
(ii) \emph{Program Accuracy}: The grounded program inferred for an (instruction, scene)-pair using the proposed model is compared with the manually annotated ground truth program. We separately report the grounding accuracy for the subject and predicate of our action (assumed binary) and the accuracy of the predicted action inferred from the instruction. Since, there is no explicit notion of grounded actions in the baseline, we do not report this metric for the baseline. 

\begin{figure}[h!]
\begin{subfigure}{0.5\hsize}
     \centering    
     \includegraphics[scale=0.19]{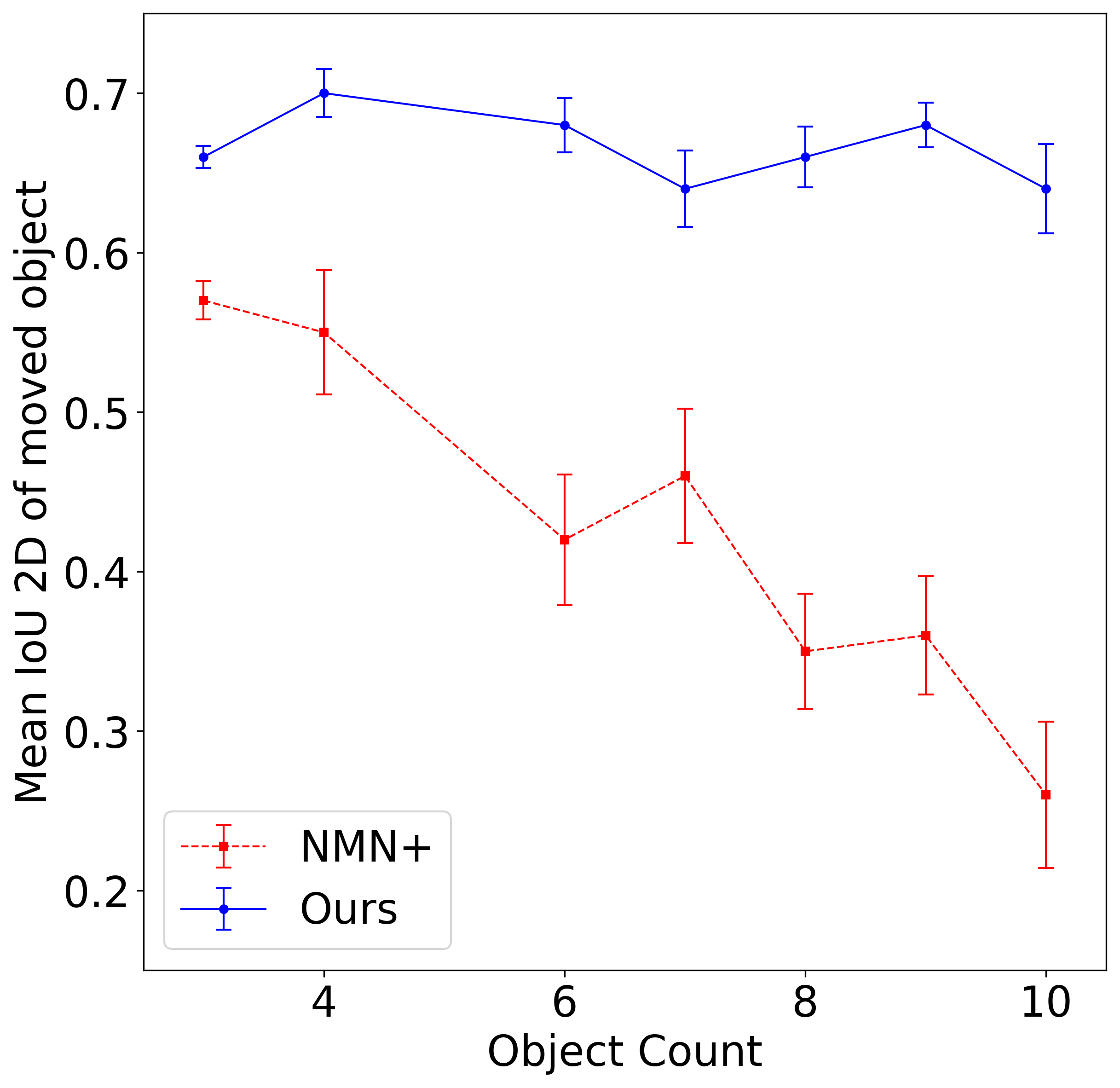}
    \caption{\footnotesize{IoU vs \# of objects}}
    \label{fig:large_scenes}
\end{subfigure}%
\begin{subfigure}{0.5\hsize}
     \centering    
    \includegraphics[scale=0.19]{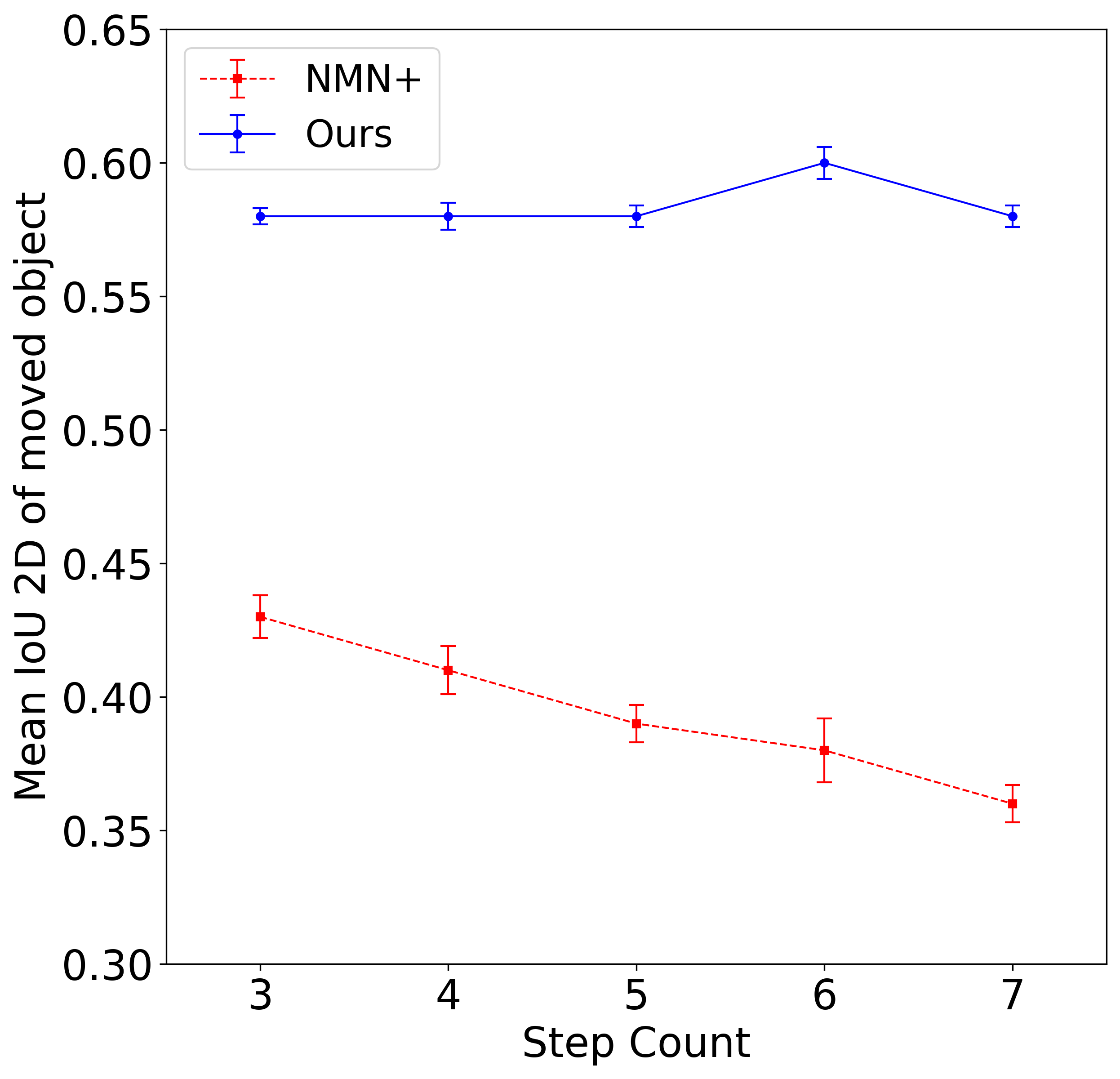}
    \caption{\footnotesize{IoU vs varying \# of steps}}
    \label{fig:large_steps}
\end{subfigure}
\caption{\footnotesize{Performance in generalization settings}}
\vspace{-0.15 in}
\end{figure}

\begin{figure*}[hbt!]
    \centering    
    \includegraphics[width=16.5cm]{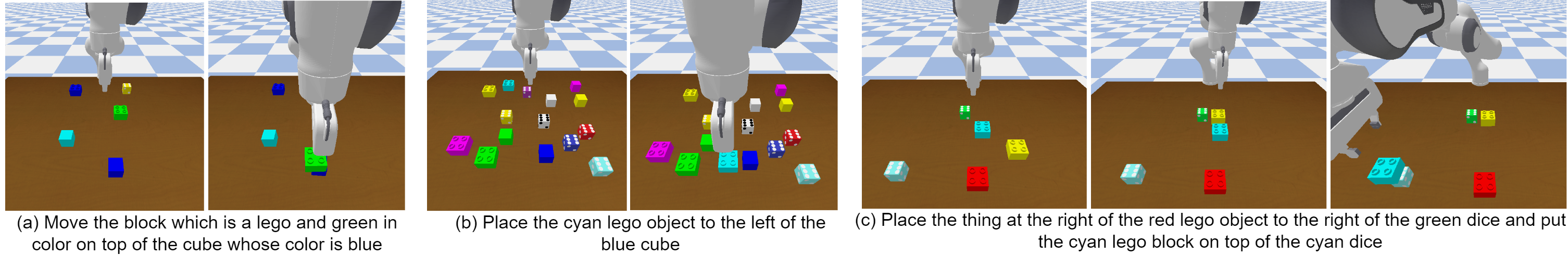}
    \label{fig:qual-1}
\end{figure*}
\begin{figure*}[hbt!]
    \centering    
    \includegraphics[width=17cm]{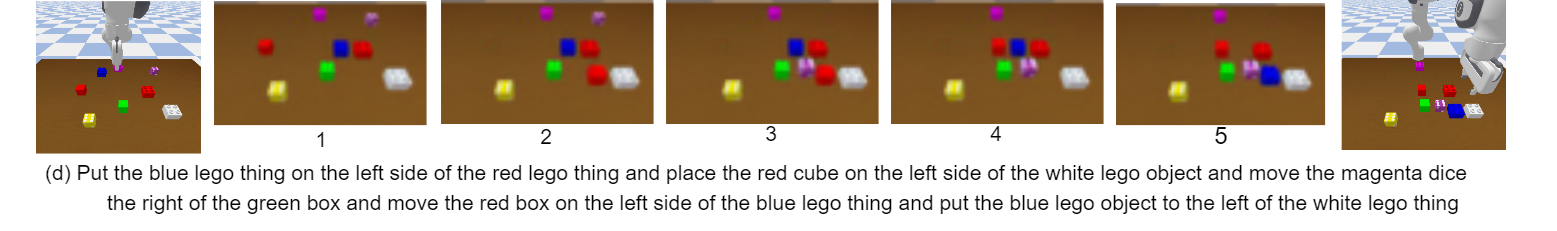}
    \caption{
    \footnotesize{
    Execution of robot manipulator on (a) compound instructions, (b) scene with 15 objects, (c) double step instruction with relational attributes, (d) 5-step instruction. (d) also shows reconstruction of the predicted scene before each step of the simulation
    }}
    \label{fig:qual-1}
    \vspace{-0.1 in}
\end{figure*}

Table \ref{tab:accuracy} reports the performance of our model and the baseline on the test set. We report both numbers: using gold set of bounding boxes and using bounding boxes extracted~\footnote{Excluded about 10\% no detection cases for both models.} using the approach described in Section~\ref{subsec:visual-reason}. Our model outperforms the baseline overall. For instructions with complex reasoning (resolution of binary spatial relations) involved, the proposed model outperforms the baseline by $33$ points in the IOU-M metric.
Disentangled representations of relations and concepts in \textit{Visual Reasoning} module allow the proposed model to reason over complex instructions.

\textbf{CLIPort}~\cite{shridhar2022cliport} takes as input a dense, image-based state representation, in contrast to our object-centric state representation. It proposes a two-stream architecture that computes language-guided attention masks on the input scene for both pick and place locations. It lacks explicit symbolic constructs for reasoning about the scene. Instead, the reasoning required for manipulation is performed by leveraging the visuo-linguistic understanding of the large pre-trained model CLIP ~\cite{clip}. Table ~\ref{tab:cliport} compares \emph{CLIPort} with our model. Since \emph{CLIPort} performs only single-step pick-and-place operations, the numbers are calculated only on single-step examples. The following metrics are used (i) \emph{Placement Accuracy}: fraction of examples in which the predicted place location lies within the groundtruth bounding box of the moved object (ii) \emph{Identification Accuracy}: fraction of examples in which the predicted pick location lies within the groundtruth bounding box of the moved object. For the proposed model, the centre of the predicted bounding box location is used as the predicted place location.

We observe that CLIPort on our dataset shows low accuracy in both predicting the object to be moved as well as the place location. Further, its performance deteriorates in examples with relational concepts. We note that the performance and sample-efficiency of CLIPort critically depends upon  spatial consistency of input images i.e. objects do not scale or distort depending on the viewing angle as in ~\cite{shridhar2022cliport}. To provide additional leverage to the baseline, we train CLIPort on 5 times more data than the original, results of which are reported under the title CLIPort-5x. We observe an improvement in performance on the larger dataset, although still worse than that of the proposed model. 
\subsection{Combinatorial Generalization} 

To evaluate the approaches in an out-of-distribution generalization setting, we collect dataset consisting of scenes with more objects than seen during training, and with instructions that involve larger number of steps for carrying out the task.

Figure~\ref{fig:large_scenes} illustrates the model generalizing combinatorially to scenes with more number of objects. The models were first trained on scenes having up to $5$ objects only, and then tested on scenes having up to $10$ objects.\\ 
The superior generalization demonstrated by the model can be attributed to reliance on an object-centric state representation and the ability to learn dense disentangled representations of spatial and action concepts, facilitating modular and structured reasoning that scales gracefully. 

Figure ~\ref{fig:large_steps} illustrates model generalization to longer horizon manipulation. We observe that the proposed model is able to perform multiple scene manipulation and reasoning steps with considerable accuracy up to $7$ steps after being trained with instructions translating to plans up to $2$ steps.\\ 
The performance of the object-centric baseline (NMN+) is worse and the model struggles to generalize to plans extending to longer horizons. We attribute this to the modular structure of our approach compared to the baseline.
%
\subsection{Demonstration on a Simulated Robot}
We demonstrate the learned model for interpreting instructions 
provided to a simulated 7-DOF Franka Emika manipulator in a table top setting. 
The robot is provided language instructions and uses the model to predict a program that once executed transitions the world state to the intended one. The 2-D bounding boxes predicted by the action simulator are translated to 3-D coordinates in the world space via a learned MLP using simulated data.
The predicted positions are provided to a low-level motion planner for trajectory generation with crane grasping for picking/placing. 
%
Figure \ref{fig:qual-1} shows execution by the robot manipulator on complex instructions, scenes having multiple objects, double step relational instructions, and multi-step instructions. We also visualise reconstruction of the moved objects before each step of the actual execution. The structural similarity index (SSIM) \cite{ssim2004} for the reconstruction model is 0.935.

%% file: conclusions.tex
\section{Conclusions}
We present a neuro-symbolic architecture that learns grounded manipulation programs via visual-linguistic reasoning for instruction understanding over a given scene, to achieve a desired goal. Unlike previous work, we do not assume any sub-goal supervision, and demonstrate how our model can be trained end-to-end. Our experiments show strong generalization to novel scenes and instructions compared to a neural-only baseline.
%
%
Directions for future work include dealing with richer instruction space including looping constructs, real-time recovery from errors caused by faulty execution, and working with real workspace data.

%% file: ack.tex
\section{Acknowledgements}
We thank anonymous reviewers for their insightful comments that helped in further improving our paper.
Parag Singla was supported by the DARPA Explainable AI (XAI) Program,
IBM AI Horizon Networks (AIHN) grant and
IBM SUR awards.
Rohan Paul acknowledges Pankaj Gupta Young Faculty Fellowship for support. 
We acknowledge the support of CSE Research Acceleration fund of IIT Delhi. Any opinions, findings, conclusions or recommendations
expressed in this paper are those of the authors and do
not necessarily reflect the views or official policies, either expressed or implied, of the funding agencies.